\documentclass{article} 
\usepackage{iclr2024_conference,times}

\usepackage{amsmath,amsfonts,bm}

\def\eqref#1{equation~\ref{#1}}

\def\1{\bm{1}}

\DeclareMathAlphabet{\mathsfit}{\encodingdefault}{\sfdefault}{m}{sl}
\SetMathAlphabet{\mathsfit}{bold}{\encodingdefault}{\sfdefault}{bx}{n}

\usepackage[utf8]{inputenc} 
\usepackage[T1]{fontenc}    
\usepackage{hyperref}       
\usepackage{url}            
\usepackage{booktabs}       
\usepackage{amsfonts}       
\usepackage{amsthm}
\usepackage{mathtools}
\usepackage{nicefrac}       
\usepackage{microtype}      
\usepackage{xcolor}         
\usepackage{color,soul}
\usepackage{graphicx}  
\usepackage{algorithm}
\usepackage{algpseudocode}
\usepackage{bbm}
\usepackage{algorithm}
\usepackage{algpseudocode}

\usepackage{amsmath,amsfonts,amssymb}
\usepackage{capt-of}
\usepackage{multirow}
\usepackage{wrapfig}
\usepackage{caption}
\usepackage{subcaption}
\usepackage{siunitx}
\usepackage{enumitem}
\usepackage{algorithm}
\usepackage{algpseudocode}
\usepackage[most]{tcolorbox}

\usepackage{hyperref}
\usepackage{url}

\definecolor{forestgreen}{rgb}{0.13, 0.55, 0.13}

\title{Enhancing Data Quality in Federated Fine-Tuning of Foundation Models}

\author{Wanru Zhao\textsuperscript{1,2}$^*$, \ Yaxin Du\textsuperscript{3}$^*$, \ Nicholas D.\ Lane\textsuperscript{1,4}, \ Siheng Chen\textsuperscript{2,3}, \ Yanfeng Wang\textsuperscript{2,3} \\
\textsuperscript{1} University of Cambridge, \textsuperscript{2} Shanghai AI Laboratory, \\ \textsuperscript{3} Shanghai Jiao Tong University, \textsuperscript{4} Flower Labs }

\iclrfinalcopy 
\begin{document}

\maketitle

\def\thefootnote{*}\footnotetext{Equal contribution.}\def\thefootnote{\arabic{footnote}}

\begin{abstract}


In the current landscape of foundation model training, there is a significant reliance on public domain data, which is nearing exhaustion according to recent research. To further scale up, it is crucial to incorporate collaboration among multiple specialized and high-quality private domain data sources. However, the challenge of training models locally without sharing private data presents numerous obstacles in data quality control. To tackle this issue, we propose a data quality control pipeline for federated fine-tuning of foundation models. This pipeline computes scores reflecting the quality of training data and determines a global threshold for a unified standard, aiming for improved global performance. Our experiments show that the proposed quality control pipeline facilitates the effectiveness and reliability of the model training, leading to better performance.

%
%


\end{abstract}

\section{Introduction}




As businesses, products, and services spring up around large language models (LLMs, which we define as having more than one billion parameters), recent work have shown that one could attain better performance by training on more high-quality data~\citep{OgScalingLaws, TrainingComputeOptimalLLMs, zhou2023lima}. However,~\citet{villalobos2022will} estimate that even high-quality English language data will be exhausted by the year 2024. What should we do when we run out of public data? 




One solution is to exploit vast private data from various institutions, including enterprises and user devices. To unlock the potential of private data, it is essential to address the following two significant issues. First, it is imperative to preserve the privacy of all participants to protect their interests~\citep{GDPR}. To address this, one could adopt federated learning~\citep{fedavg}, a collaborative machine learning framework that trains a model across multiple clients with their local private data, without exchanging any raw data.
Second, enhancing model training requires the control the data quality from each participant. Due to the inability to directly access private data, quality control for these data poses significant challenges, which is the focus of this work.



Previously, data quality control relied heavily on manual selection processes~\citep{llama2, llama}. This approach, while commonly used, presented significant challenges due to the high volume of data, leading to substantial costs. Recent advancements have seen the introduction of automated low-quality data filters~\citep{together2023redpajama}, such as perplexity filters~\citep{muennighoff2023scaling} and deduplication filters~\citep{lee2021deduplicating}. These automated methods are designed to reduce data volume and enhance training efficiency in centralized settings, while their effectiveness in data quality control within collaborative environments remains to be explored.




In our paper, 
we propose an automated data quality control pipeline for federated fine-tuning of large language models (LLMs), showcasing notable performance improvements in mixed-quality data environments. Specifically, we incorporate data valuation algorithms to serve as scoring functions, enabling fine-grained evaluation of individual training sample quality. Furthermore, we establish a unified data quality standard using a minimal set of anchor data, addressing the challenge of heterogeneity in data quality across federated institutions. 
Adopting this approach, we effectively eliminate low-quality data, thereby enhancing model performance and ensuring privacy preservation. Leveraging the collaboration of multiple private domain data sources, opens up new possibilities in the face of real-world public data exhaustion.


\section{Motivation and Setup: How low-quality data affects the performance of collaborative training} \label{sec:motivation}

In our paper, we identify two unique challenges for  federated fine-tuning of LLMs in terms of data quality. 1) \textbf{Real low-quality data}
Firstly, we aim to highlight three prevalent patterns of low-quality data observed in real-world corpora: cut, deletion and exchange. The cut category encompasses scenarios where content is truncated due to word limit constraints, deletion pertains to instances where critical terminologies are absent from the corpus, and exchange refers to examples containing entirely incorrect information. We provide specific examples of these categories in Appendix~\ref{app:low_quality_making}.
\begin{wrapfigure}{r}{0.5\linewidth}
  \centering
  \vspace{-10pt}
  \includegraphics[width=\linewidth]{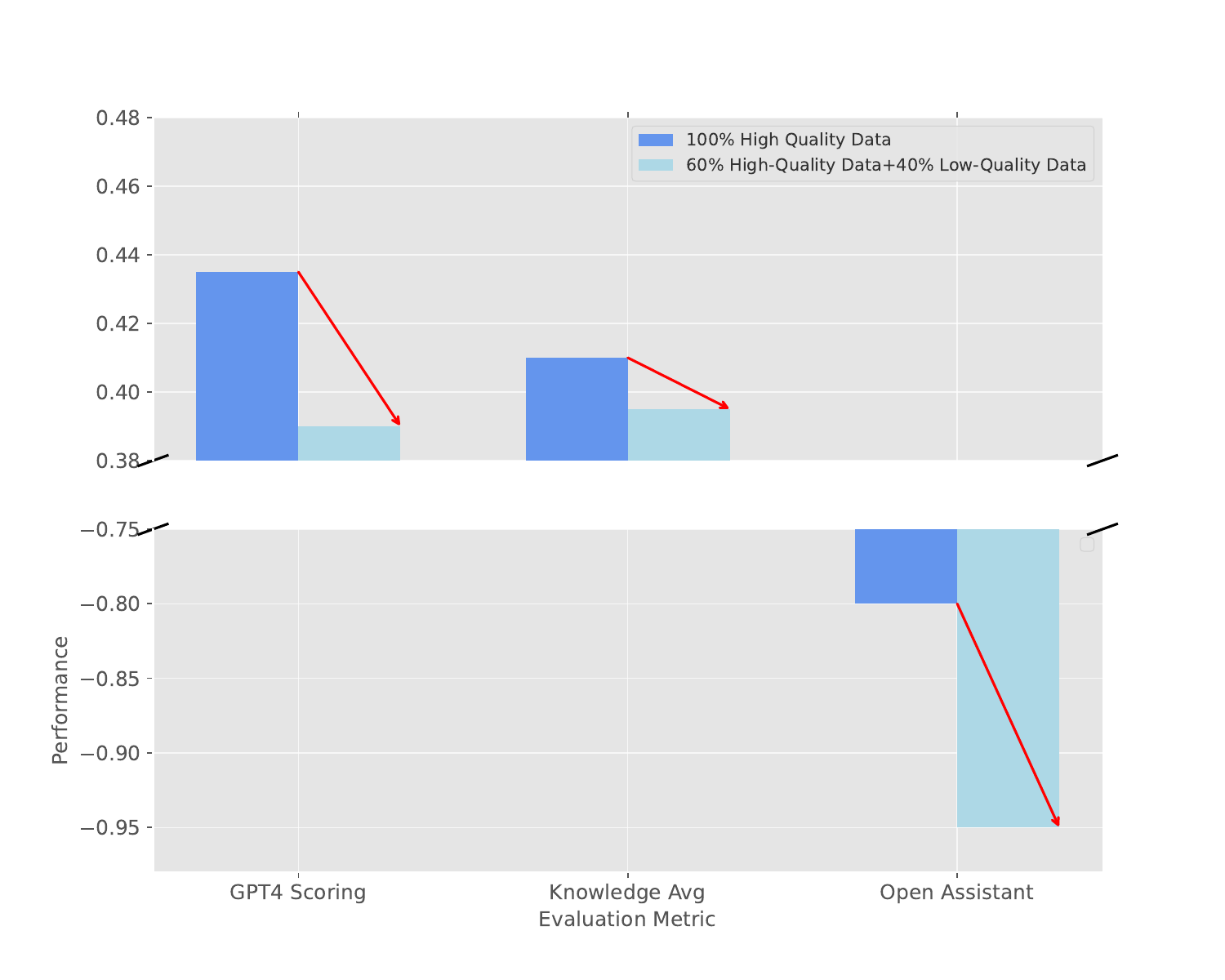} 
  \caption{The impact of low-quality data on the performance of federated fine-tuning of LLMs .
  }
  \label{fig:motivation}
  \vspace{-10pt}
\end{wrapfigure}
2) \textbf{Quality heterogeneity}
Quality heterogeneity refers to the variability in the quality of data collection across different clients in federated learning. Given that federated learning often encompasses a vast number of clients, each with varying capabilities in data synthesis, it is impractical to assume uniformity in data quality among all participants. Consequently, some clients may possess a higher proportion of low-quality data compared to others, highlighting the absence of a uniform standard for data sample quality across all participants. We provide two Non-IID settings in Appendix~\ref{app:NIID}.

In our preliminary experiments, we consider the two factors above, adjust the proportion of low-quality data of composition of PMC-LLama~\citep{wu2023pmc} and Medalpaca-flashcards~\citep{han2023medalpaca} datasets in federated training, shown in Figure~\ref{fig:motivation}. Higher scores indicate better performance (for more details about the metrics, see Appendix~\ref{app:metrics}). The key observation is, the quality of the training data has a significant effect on the performance of collaborative training: low-quality data consistently lead to worse influence on all the metrics.




\vspace{-5pt}
\section{Proposed Workflow for Data Quality Control}


\subsection{Overview}
\begin{figure}[t]
  \centering
  \vspace{-20pt}
    \begin{subfigure}{0.49\textwidth}
        \centering
        \includegraphics[width=\textwidth]{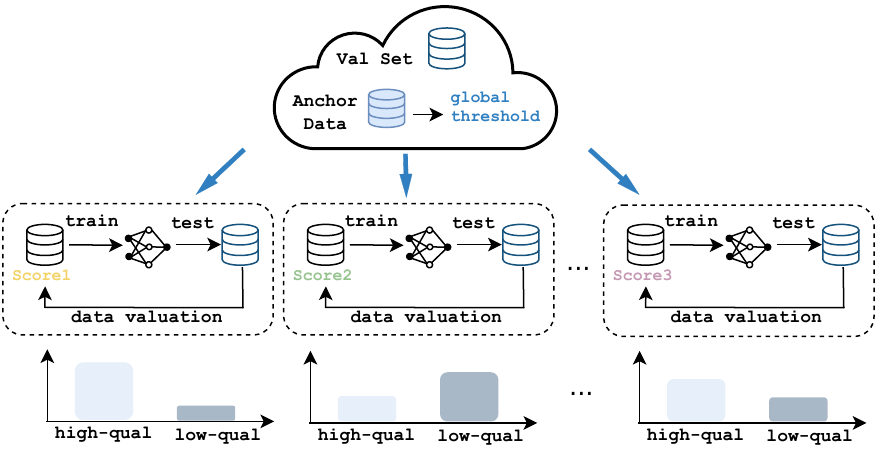}
        \caption{Phase I: Pre-training: Data Quality Control}
    \end{subfigure}
    \hfill
    \begin{subfigure}{0.49\textwidth}
        \centering
        \includegraphics[width=\textwidth]{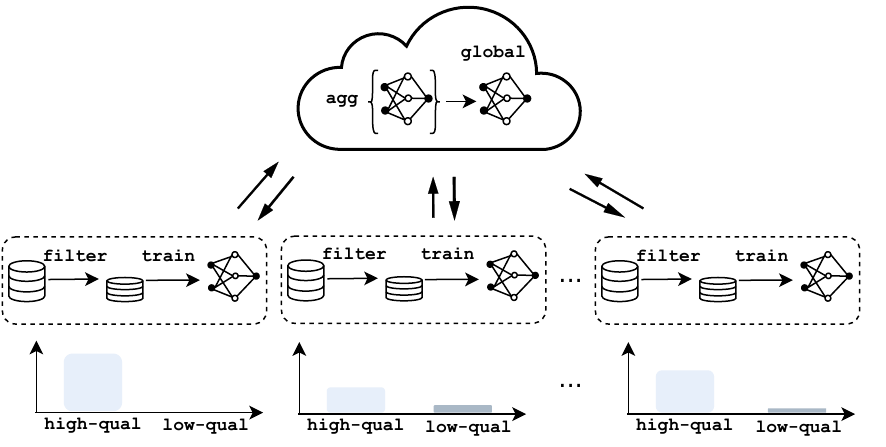}
        \caption{Phase II. FL with High-Quality Data}
    \end{subfigure}
  \hfill
  \caption{Overall workflow diagram consists of two phases: 1) Phase I: client-side compute each sample’s quality score with scoring functions using the public validation set and global model, then server-side aggregates the scores, giving a global threshold by anchor data 2) Phase II: clients filter data according to the global threshold and starts federated learning on selected high-quality data.}
  \label{fig:mainfig}
\end{figure}

Federated learning workflow iterates two steps on the client and server side: 1) client downloads global model $\bm{\theta}$ from the server and conducts training on local dataset $\mathcal{D}_k$ to obtain local model $\bm{\theta}_k$; 2) server receives local models $\{\bm{\theta}_k\}$ from clients and aggregates them to obtain global model $\bm{\theta}:=\sum_k p_k \bm{\theta}_k$.
In our workflow, each client conducts local training using its private, high-quality data. We have a public validation set located on both the clients and the server, which consists of commonly recognized, high-quality public data. As illustrated in Figure~\ref{fig:mainfig}, the overall workflow consists of two phases designed to achieve data quality control in the federated training of LLMs. 

\textbf{Phase I. Pre-training: Data Quality Control} 
Client receives initial global model $\theta^0$ and public validation set $\mathcal{D}_{val}$ from server. Then clients compute each sample's quality score with scoring functions using the public validation set and global initial model. The server determines a global threshold of score, serving as a unified standard of data quality with only a very few amount of anchor data, and sends it to the clients.  

\textbf{Phase II. Federated Learning with High-Quality Data} Each client then discards data samples that fall below the received global threshold, ensuring that only high-quality data verified by the unified standard is retained. Then clients utilize the filtered high-quality data sets $\mathcal{D}_k'$ (where $|\mathcal{D}_k'|\leq|\mathcal{D}_k|$) and global inital model $\theta^0$ for federated learning.


\subsection{Local Data Scoring and quality control}


For each client, we have designed a data scoring step in which we calculate the score for each training data sample to assess its contribution to model prediction. Specifically, considering an ordered training set of examples \(\mathcal{D}_{k} = \{z_1, \ldots, z_n\}\) and a model \(\theta\), we define the scoring function \(S(z_i, \theta)\) for each data sample. Those with low scores are filtered out in subsequent steps. Following the inherent principle, we incorporate the concepts of existing measurements into our scoring method, which includes probability-based scoring and gradient-based scoring.

\paragraph{Perplexity:}
As a probability-based scoring method, perplexity is defined as the exponentiated average negative log-likelihood of a sequence. For a tokenized sequence $X=\left(x_0, x_1, \ldots, x_t\right)$, the perplexity scoring function of $X$ is, $
\mathrm{PPL}(X)=\exp \left\{-\sum_i^t \log p_\theta\left(x_i \mid x_{<i}\right)/t\right\},
$
wherelog $p_\theta\left(x_i \mid x_{<i}\right)$ is the log-likelihood of the ith token conditioned on the preceding tokens $x_{<i}$ according to our model. It is commonly used as an evaluation of the model's ability to predict uniformly among the set of specified tokens in a corpus. Here it can be used as the scoring function for each individual data sample. 
\vspace{-10pt}
\paragraph{Conditional Probability:} 
This is another probability-based scoring method. Each word in Question $(Q)$ and Answer $(A)$ is denoted as $x_i^Q$ and $x_i^A$ respectively. Conditional Probability on following instruction of a given $(Q, A)$ pairs by calculating the ratio between $Pro_\theta(A)$ and $Pro_\theta(A \mid Q)$ :
$
\text{Prob}(X) = \sum_{i} \log p_{\theta}(x_i | x_{<i})/t,
\text{ConProb}(Q, A) = {\text{Prob}(A | Q)} / {\text{Prob}(A)}.
$
Conditional probability scores demonstrate how well the model aligns each response to the given corresponding question.

\paragraph{Influence Functions:}
This is a gradient-based scoring method based on the validation performance. The standard Hessian-based influence functions yield scores $
\operatorname{DataInf}\left(x_j\right)_i=\nabla L\left(x_j ; \theta^{\star}\right) H_{\theta^{\star}}^{-1} \nabla L\left(x_i ; \theta^{\star}\right)
$ for all $x_i\in\mathcal{D}_k$ and $x_j\in\mathcal{D}_{val}$, 
where $\theta^{\star}$ is the model trained on train set and $H_{\theta^{\star}}$ is the Hessian of the empirical loss. By fat, DataInf~\citep{kwon2023datainf} is the first computationally efficient influence approximation method that can be easily applied to LLM.

\subsection{Global Standard with Anchor Data Scoring} \label{anchor_data} 


On the server, we select only a few amount of data (10 samples in our paper) as our anchor data and use the aforementioned scoring method to calculate the average score of these 10 data points as the global threshold. This establishes a unified standard for division between low- and high-quality data for heterogeneous clients, allowing for the further filtering of local data.

\section{Experiments}
\subsection{Experiment Setup}
\paragraph{Tasks and Datasets}
We conduct our experiments on the question-answering~(QA) task, sampled from the following two datasets. \textbf{1) PMC-LLama}~\citep{wu2023pmc} contributes a large-scale, comprehensive dataset for instruction tuning. This dataset encompasses medical question-answering, rationale for reasoning, and conversational dialogues, comprising a total of 202M tokens. \textbf{2) Medalpaca-flashcards}~\citep{han2023medalpaca} presents an innovative dataset consisting of over 514k entries, specifically crafted to fine-tune LLMs for effective medical applications. 

\vspace{-5pt}
\paragraph{Models}
Our pre-trained base model is the LLama2-7b model~\citep{llama2}, shown to perform well across many domains. We adapt commonly-used parameter-efficient fine-tuning technique, Low-Rank Adaptation~(LoRA), which involves freezing the pre-trained model weights and injecting trainable rank decomposition matrices into each layer of the Transformer architecture, thereby achieving fine-tuning without incurring any additional inference latency~\citep{lora}. 

\vspace{-5pt}
\paragraph{Evaluation metrics}
In this study, we classify three evaluation metrics across five benchmark tests commonly utilized in contemporary medical LLMs. ~\citep{medpalm2, wu2023pmc, lee2020biobert, luo2023biomedgpt} The evaluations focus on two main aspects: 1) \textbf{question-answering capabilities}, assessed by GPT-4 ~\citep{openai2023gpt4} and Openassistant ~\citep{kopf2023openassistant} scoring within the Medalpaca-flashcards test set. The former metric gauges the consistency of the generated responses with the ground truth, while the latter simulates the human reward of the answers. 2) \textbf{Knowledge acquisition ability}, measured by average accuracy of responses to multiple-choice questions in the MMLU~\citep{hendrycks2021ethics}, MedMCQA~\citep{medmcqa}, PubMedQA~\citep{pubmedqa}, and USMLE~\citep{usmle} datasets. 


\subsection{Experimental Results}


Based on the low-quality dataset setup described in the Section~\ref{sec:motivation}, we build up our our data-quality control pipeline in centralized, federated IID and Non-IID setting (described in Appendix \ref{app:fedsetting}). We implement both probability-based and gradient-based scoring methods to calculate scores for each training data, and set the unified scoring standard using corresponding scoring functions with anchor data. In comparison, we add a baseline, \textit{Oracle}, which only trains on high-quality data. We also consider the commonly used in-context learning (ICL) as another scoring method, provided in Appendix~\ref{app:benchmark}.

\begin{table}[ht!]
\centering
\label{tab:main}
    \vspace{-5pt}
    \caption{Comparison of five data quality control methods across centralized, federated IID, and federated Non-IID settings with three evaluation metrics.}
    \renewcommand{\arraystretch}{1.1}
    \vspace{-5pt}
    \scalebox{0.75}{
\begin{tabular}{c|ccc|ccc|ccc}
\toprule
         & \multicolumn{3}{c}{Centralized ($n=1$)}              & \multicolumn{3}{c}{Federated~($n=20$, IID)}                      & \multicolumn{3}{c}{Federated~($n=20$, Non-IID)}                     \\ \midrule
         \begin{tabular}[c]{@{}c@{}}Evaluation\\ Metric\end{tabular} & \begin{tabular}[c]{@{}l@{}}GPT-4 \\ Scoring\end{tabular} & \begin{tabular}[c]{@{}l@{}}Open\\ Assistant\end{tabular} & \begin{tabular}[c]{@{}l@{}}Knowledge\\ Avg\end{tabular} & \begin{tabular}[c]{@{}l@{}}GPT-4 \\ Scoring\end{tabular} & \begin{tabular}[c]{@{}l@{}}Open\\ Assistant\end{tabular} & \begin{tabular}[c]{@{}l@{}}Knowledge\\ Avg\end{tabular} & \begin{tabular}[c]{@{}l@{}}GPT-4 \\ Scoring\end{tabular} & \begin{tabular}[c]{@{}l@{}}Open\\ Assistant\end{tabular} & \begin{tabular}[c]{@{}l@{}}Knowledge\\ Avg\end{tabular} \\ \midrule
Low-qual Data & 0.389         & -0.89         & 0.398        & 0.411         & -0.83         & 0.348        & 0.392         & -0.87         & 0.396        \\

Oracle   & 0.436         & -0.63         & 0.442        & 0.455         & -0.79         & 0.440        & 0.401         & -0.75         & 0.417        \\
\hline
PPL      & \textbf{0.430}         & \textbf{-0.84}         & \textbf{0.399}        & 0.392         & -0.92         & \textbf{0.362}        & \textbf{0.414}         & -0.91         & 0.374        \\
ConPro   & \textbf{0.441}         & \textbf{-0.75}         & \textbf{0.445 }       & \textbf{0.420}         & -0.86         & \textbf{0.443 }       & \textbf{0.419}         & -0.87         & \textbf{0.444}        \\
DataInf  &       0.381        &     \textbf{-0.74}          &         \textbf{0.412}     & 0.411         & \textbf{-0.47}         & \textbf{0.373}        & \textbf{0.398}         & \textbf{-0.66}         & \textbf{0.423}        \\
ICL      & \textbf{0.438}         & \textbf{-0.68}         & \textbf{0.446}        & \textbf{0.418}         & \textbf{-0.86}         & \textbf{0.437}        & \textbf{0.435 }        & \textbf{-0.84}         & \textbf{0.448}  \\
\bottomrule

\end{tabular}
\vspace{-30pt}
}

\end{table}


The experimental results are presented in Table 1. Our findings include: \textbf{1)} Utilizing our current data quality control pipeline, our data scoring techniques consistently outperform the models trained on low-quality datasets in both centralized and federated settings. \textbf{2)} When employing \textit{ConPro} and \textit{ICL} as data scoring methods to select high-quality data, the performance of the global model can surpass that of models trained with the oracle set. \textbf{3)} In the Non-IID setting, the global model trained with our quality-controlled data demonstrates excellent performance. These experimental results highlight not only the efficacy of the data scoring methods but also the effectiveness of a unified quality threshold determined by anchor data (for more details, see Appendix~\ref{app:ablation}).





\section*{Conclusion and Future Work} 



In this paper, we establish a data quality control pipeline for federated fine-tuning of LLMs, avoiding directly sharing any private data. Preliminary experiments show that the selected high-quality data ensures an effective and reliable learning process, leading to improved model performance. To further safeguard privacy, we can seamlessly integrate differential privacy or a secure aggregation component to prevent reconstruction attacks, enhancing the security of our framework.





\clearpage
\bibliography{iclr2024_conference}
\bibliographystyle{iclr2024_conference}

\clearpage
\appendix

\section{Related Work}
\paragraph{Large Language Models} The remarkable achievements of large language models (LLM) have recently  impacted the field of natural language processing. OpenAI’s GPT-4~\citep{openai2023gpt4}, for instance, has demonstrated exceptional capabilities in various generative tasks including question answering. LLaMA~\citep{llama, llama2}, an open-source large language modelwith 7 to 65 billion parameters, offers an alternative platform for research.  These developments have sparked interest in adapting LLMs for medical applications. Yet, most medical models are fine-tuned based on LLaMA on a small medical corpus, resulting in a deficiency of comprehensive medical knowledge integration. There has been recent efforts on training LLMs for medical domains, for example, BioBert~\citep{lee2020biobert}, BioMedGPT~\citep{luo2023biomedgpt}, PMC-LLama~\citep{wu2023pmc}. These domain-specific LLMs have been exclusively trained on medical corpora. However, there has been a lack of collaborative or federated training work in medical LLMs.

\paragraph{Data Valuation and Attribution} 
The seminal work on data valuation/attribution of~\citet{koh2017understanding} proposes attribution via approximate influence functions. It identifies training samples most responsible for a given prediction by estimating the effect of removing or slightly modifying a single training sample. 
In a related approach, TracIn~\citep{tracin} estimates the influence of each sample in training set on the test example by measuring the change in loss from gradient updates of mini-batches. 
Another related line of work has utilized Shapley values~\citep{lundberg2017unified} to ascribe value to data, but Shapley values often require exponential time to compute. 

\paragraph{Federated Learning} Federated Learning has garnered significant attention as a distributed machine learning paradigm. It shifts the traditional model training process by sharing model parameters instead of raw data. With Federated Averaging~(FedAvg)~\citep{fedavg}, participating clients train models using their own private datasets locally, and the updated model parameters are aggregated on the server. This preserves the privacy of the underlying data while collectively benefiting from the knowledge gained during the training process~\citep{konevcny2016federated2}. Recent work~\citep{FedPromptTuning} proposes a federated parameter-efficient fine-tuning paradigm for large language models, demonstrating not only its advantages in data- and parameter-efficiency but also in better generalization and stability.
Despite abundant research made on federated fine-tuning LLMs, data quality control in federated manner remains under-explored.


\section{Heterogeneity Settings}\label{app:NIID}

\begin{figure}[t]
  \centering
    \begin{subfigure}{0.49\textwidth}
        \centering
        \includegraphics[width=\textwidth]{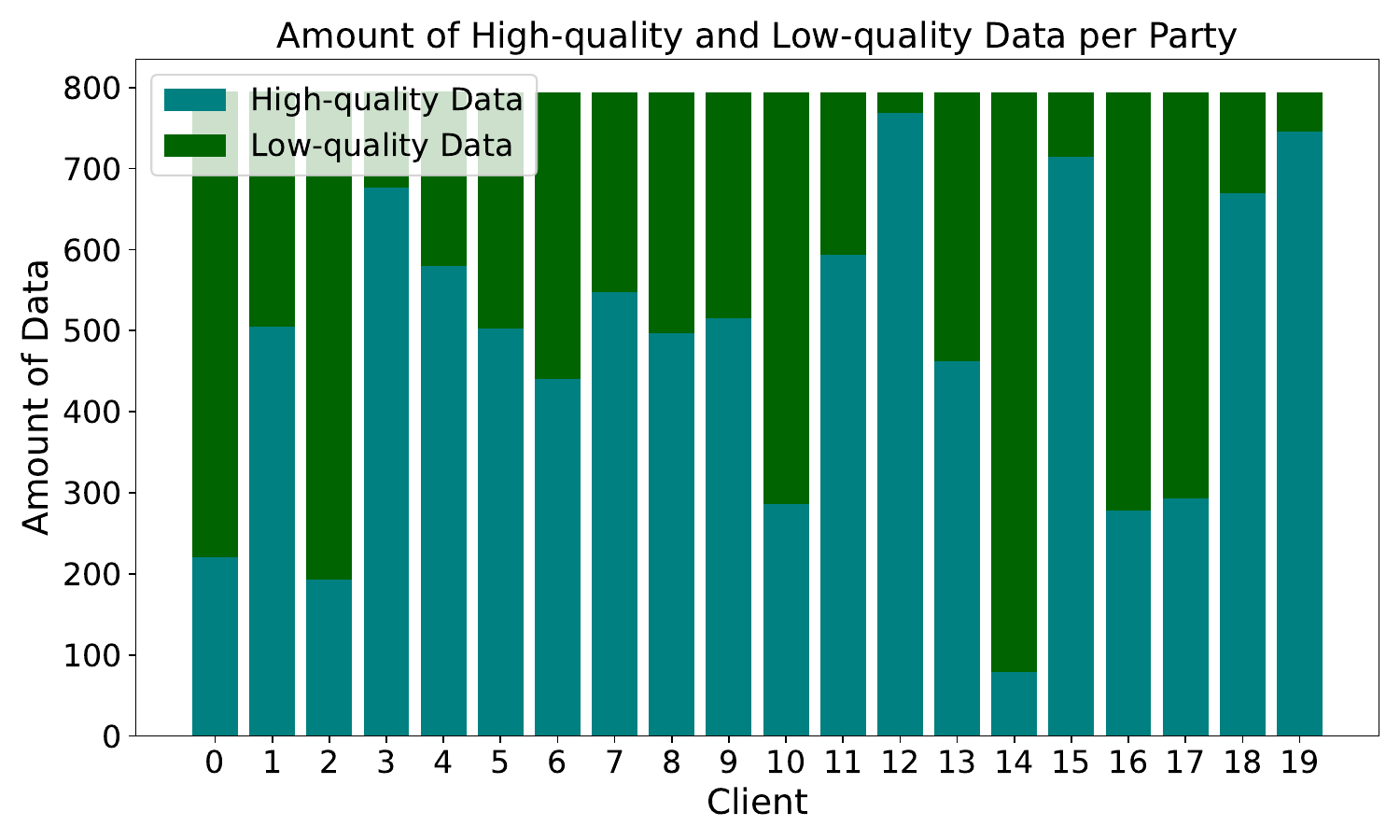}
        \caption{NIID-1}
    \end{subfigure}
    \hfill
    \begin{subfigure}{0.49\textwidth}
        \centering
        \includegraphics[width=\textwidth]{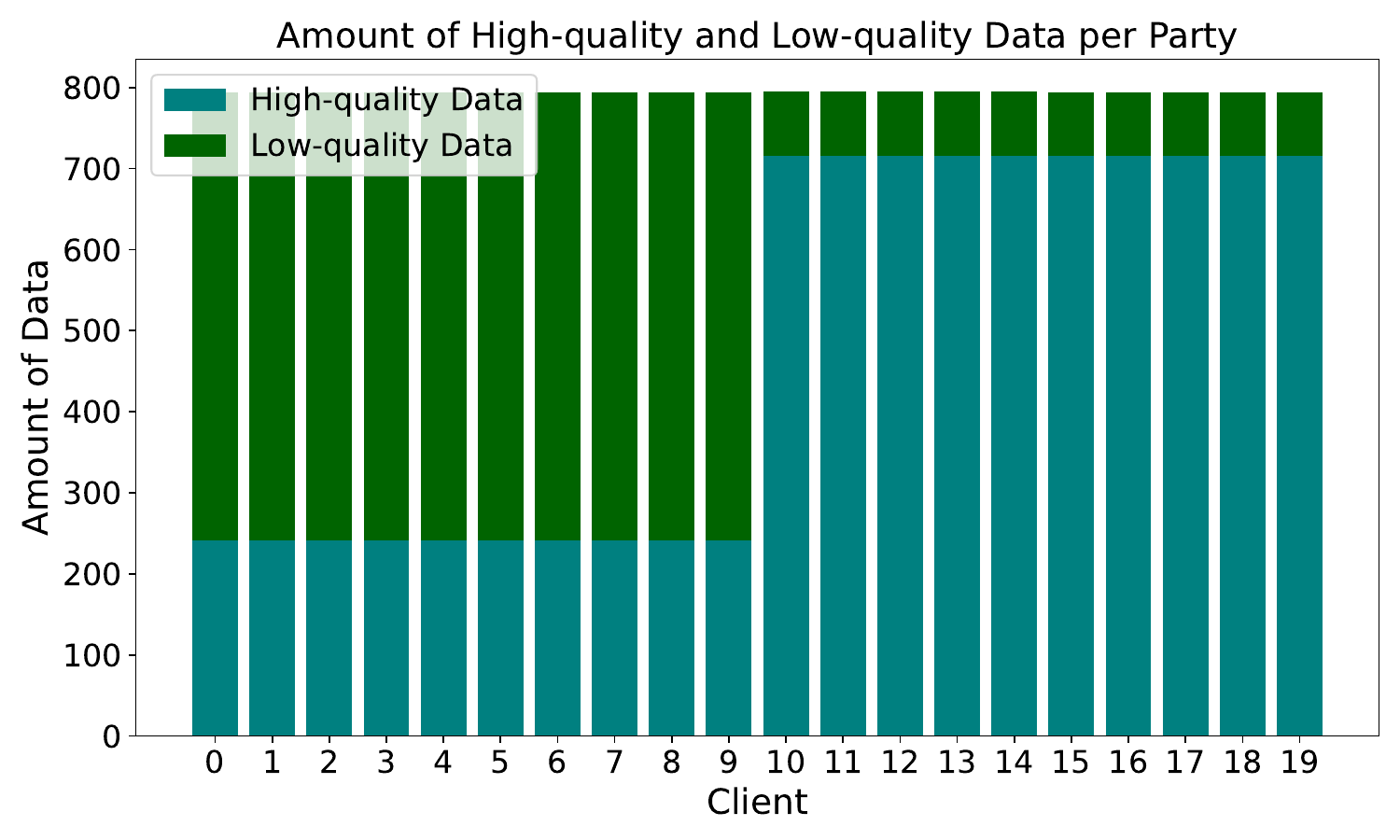}
        \caption{NIID-2}
    \end{subfigure}
  \hfill

  \caption{Data compromisation of high-quality and low-quality data with NIID-1 and NIID-2}
  \label{fig:distribution}
\end{figure}

To model real-world scenario, we designed two heterogeneous settings: NIID-1 and NIID-2. NIID-1 replicates a typical scenario in federated learning classification tasks~\citep{bayesian, fedma, fednova, moon, feddecorr}, where the distribution of low-quality data among clients follows a Dirichlet distribution with parameter $\beta=1$, while ensuring that the volume of data processed by each client remains equal. In contrast, NIID-2 addresses a skewed classification task scenario within FL~\citep{fedavg, fedprox}, assigning $70\%$ of low-quality data to half of the clients and $90\%$ to the other half, yet maintaining an equal size of training data across all clients. The distributions for these settings are illustrated in Figure~\ref{fig:distribution}. Table\ref{tab:niid2} shows the low-quality data traing and data quality control federated NIID-2 setting.

\begin{table*}[ht!]
    \caption{Comparison of four data quality control methods on
federated NIID-2 settings with three evaluation metrics.}
    \vspace{1mm}
    \label{tab:niid2}
    \centering
    \renewcommand{\arraystretch}{1.5}
    \scalebox{0.7}{
    \begin{tabular}{lc c c c}
    \toprule
    \multicolumn{2}{l}{Setting} & \multicolumn{1}{c}{GPT-4 Scoring} & \multicolumn{1}{c}{OpenAssistant} & \multicolumn{1}{c}{KnowledgeAvg} \\
    \hline

    \multirow{4}{*}{$n=20$, NIID-2} & Federated Low-qual Data & 0.399 & -0.94 & 0.399  \\
    
    & Oracle & 0.418  & -0.86 & 0.413 \\
    \cline{2-5}
    & ConPro & \textbf{0.427} & \textbf{-0.78} & \textbf{0.414} \\
    & PPL & 0.389 &  \textbf{-0.87} & 0.199 \\
    & DataInf & 0.381  & -0.98 & \textbf{0.424} \\
    & ICL & 0.341 &  \textbf{-0.85} & \textbf{0.426} \\
    \bottomrule
   \end{tabular}  
}
\end{table*}

\section{Experimental Details}

\subsection{Implementation Details}

\paragraph{Training setups}
\label{app:fedsetting}
We use Hugging Face's transformers library~\citep{huggingface-transformers} and PEFT library~\citep{peft} for loading pre-trained models and LoRA configurations. For our training and evaluation, we use PyTorch as the underlying auto-differentiation framework~\citep{pytorch}. We use the AdamW optimizer ~\citep{adamw, adam} for all experiments. All the experiments are conducted on the machines with the same hardware configuration using NVIDIA A40. 
In all experiments, we use 8 bit quantization, set $batch\_size=16$, and LoRA rank to 16. For the federated setting, we consider 300 communication rounds, $n=20$ clients, 10 local steps of model training (equal to 3 epoch for local training).

\paragraph{Training data}
\label{app:dataset_making}
We use 16k sample in total, with 8k samples randomly sampled from PMC-LLama~\citep{wu2023pmc} and Medalpaca-flashcards~\citep{han2023medalpaca} each. In the low-quality data synthetic processes, 3.2k samples (40\% total data) are either polluted with cut (10\% total data), delete (15\% total data) and exchange (15\% of total data) tricks. These 40\% low-quality data together with the rest of high-quality data composites the low-quality data set. While the oracle data set only includes the rest of high-quality data set.


\subsection{Evalution metrics} \label{app:metrics}

\paragraph{GPT-4 scoring} 

200 samples are randomly selected from the Medalpaca-flashcards dataset to serve as the test set. We evaluate the models that need to be compared on the test set to generate responses respectively. Then we use OpenAI GPT-4 model API to assign scores for their responses. Each response of is rated by the judge on a scale from 0 to 1, reflecting how well the answer aligns with the ground truth. 

\paragraph{OpenAssistant} 
OpenAssistant~\citep{kopf2023openassistant} is a reward model built on the DeBERTa~\citep{he2020deberta} architecture, trained with human feedback data from studies by~\citet{nakano2021webgpt, stiennon2020learning, bai2022training}. 
In this metric, 200 samples were randomly selected from the Medalpaca-flashcards dataset as the test set, with each assigned a score from -5 to 5 to reflect data quality.

\paragraph{Knowledge-based benchmarks}
For evaluation on multiple-choice questions, we use the USMLE~\citep{usmle}, MedMCQA~\citep{medmcqa}, PubMedQA~\citep{pubmedqa} and MMLU clinical topics~\citep{hendryckstest2021} datasets. More details refer to Appendix~\ref{app:benchmark}.

\subsection{In Context Learning}\label{app:ICL}
In-context learning primarily denotes the capacity of LLMs to adapt to new information or tasks based solely on the context presented in their input, eliminating the necessity for explicit retraining or fine-tuning for those specific tasks. Meanwhile, some work~\citep{zhou2023lima} hypothesizes that most knowledge in LLM is learned from the pre-training stage, while instruction tuning stimulates the instruction-following ability. Therefore, in the realm of data selection, we assume pre-trained LLM is able to select obvious low-quality data by prompting pre-trained model with serveral examples. Sepicifally, incontext learning method provides several examples $(\hat{Q}, \hat{A})$ to a pre-trained model with respect to specific sample $(Q, A)$. we form the input as $\overline{Q}=[(\hat{Q}, \hat{A}), Q]$, and take the in-context learning score as $l(A,p_{\theta}(A|\overline{Q}))$, where $l$ is the training loss.

\subsection{Knowledge-based benchmark}\label{app:benchmark}
\paragraph{MMLU} MMLU~\citep{hendryckstest2021} provides a comprehensive suite of tests for assessing text models in multi-task contexts. 
We utilized the clinical topics from the MMLU test set as our testing ground, encompassing a diverse range of subjects including 265 questions on Clinical Knowledge, 100 on Medical Genetics, 135 on Anatomy, 272 on Professional Medicine, 144 on College Biology, and 173 on College Medicine, all formatted as multiple-choice questions.

\paragraph{MedMCQA} 
MedMCQA~\citep{medmcqa} is a dataset comprised of multiple-choice questions derived from mock exams and past papers of two major Indian medical school entrance examinations, namely AIIMS and NEET-PG. The dataset is divided into two parts: the training split, which includes 182,822 questions, and the test split, consisting of 4,183 questions. Each question in the dataset is accompanied by four possible answers.

\paragraph{PubMedQA}
PubMedQA~\citep{pubmedqa} is a biomedical QA benchmark collected from PubMed abstracts. The PubMedQA task is designed to answer research questions with responses categorized as yes/no/maybe, effectively framing it as a multiple-choice question format. The dataset is divided into three subsets: 1,000 manually labeled question-answer pairs (denoted as PQA-L), 61,200 unlabeled pairs (PQA-U), and 211,300 pairs that have been artificially generated (PQA-A). Consistent with previous studies~\citep{lmflow, medpalm2}, we employ the PQA-L subset as the test set for evaluating the model's performance.

\paragraph{USMLE}
USMLE~\citep{usmle} consists of multiple-choice questions (with 4 choices per question) that are based on the United States Medical Licensing Exams. This dataset has been compiled from questions used in professional medical board examinations and is unique in its multilingual composition, including English, Simplified Chinese, and Traditional Chinese versions. It contains 12,724 questions in English, 34,251 in Simplified Chinese, and 14,123 in Traditional Chinese. For our purposes, we focus on the English component of the dataset, which is further divided into 10,178 questions for the training set, 1,273 for the validation set, and 1,273 for the test set, adhering to the official distribution of the dataset.

\section{Ablation study of unified scoring with anchor data} \label{app:ablation}

\begin{figure}[htb]
    \centering
    \includegraphics[width=0.5\linewidth]{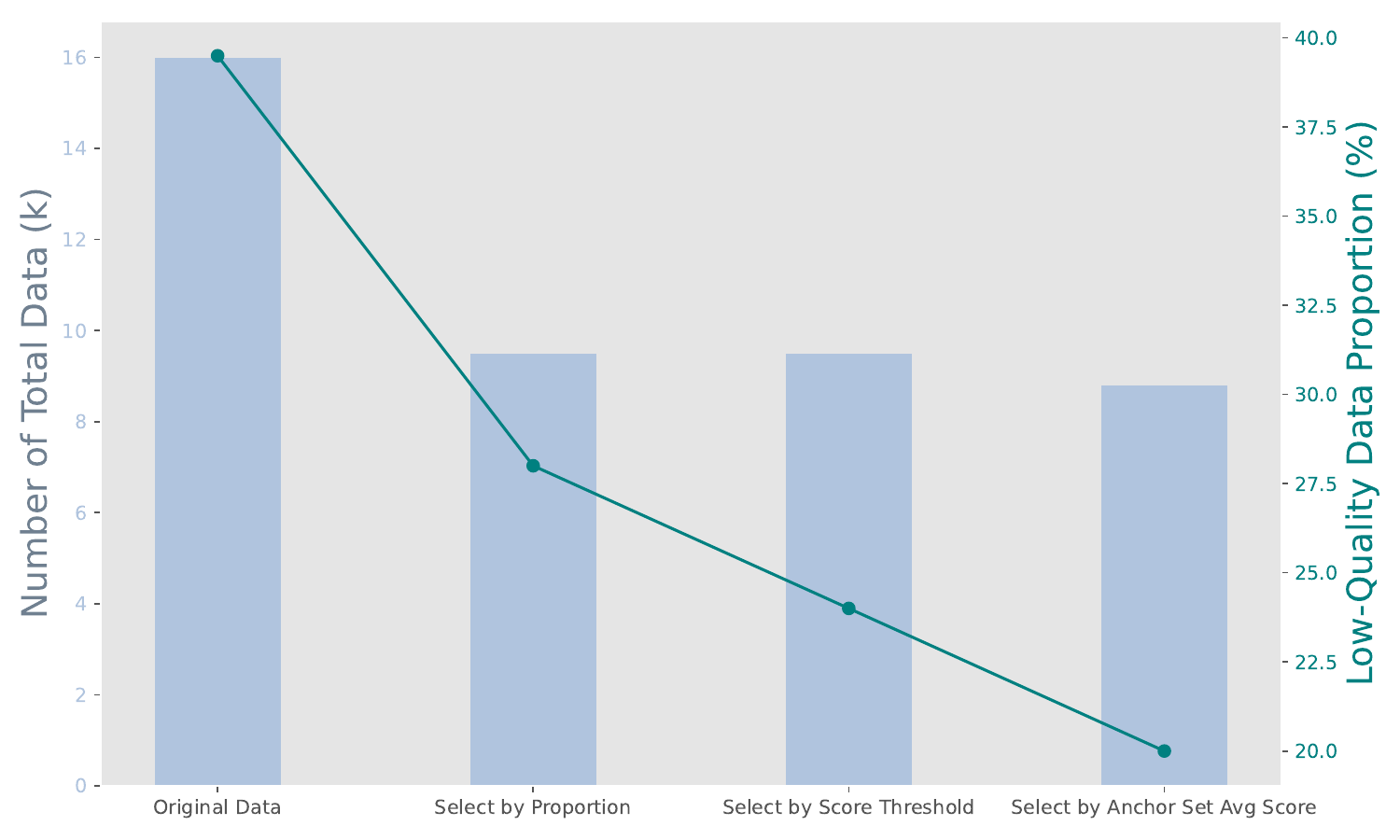}
    \caption{ Number of selected data and the proportion of low-quality data across different selection principles (select by proportion, score threshold, or anchor set score) in federated NIID1 setting, employing the ConPro score.}
    \label{fig:unified_score}
\end{figure}

We perform a comparative analysis focusing on the number of selected data and the proportion of low-quality data. This comparison contrasts the approach of selection based on a unified scoring from anchor data, as depicted in the fourth bar, with methods that determine the threshold by acknowledging the overall proportion of low-quality data. In Figure \ref{fig:unified_score}, the method represented by the second bar involves each client selecting high-quality data by sorting each sample's scores and eliminating the lowest-scored samples, informed by the proportion of low-quality data within the entire training set of FL. This approach, which does not need a score threshold and relies solely on the known proportion of low-quality data, proves to be impractical and less effective in settings with heterogeneous data quality due to varying data quality compositions among clients. The third bar illustrates the outcomes of implementing a global score threshold aligned with the exact proportion of global low-quality data. This method necessitates the server's aggregation of all scores from clients within the FL framework and knowing the proportion of low-quality data, potentially leading to privacy concerns. The results indicate that selection by anchor score consistently yields a lower proportion of low-quality data compared to the other two methodologies.

\section{Examples for low- and high- quality data} \label{app:low_quality_making}

To model real-world scenario, we designed three types of low-quality data generation rules. 

\paragraph{Cut} The cut type data simulates scenarios where an answer exceeds the word limit, resulting in the truncation of the last few words. Practically, we establish a threshold of 100 tokens, retaining only the initial tokens up to this limit, as demonstrated in Table \ref{tab:cut_example}.

\paragraph{Deletion} Deletion type of data refers to missing important terminologies. In practice, we delete part of the tokens in the answer to simulate the missing terminology or some nouns, shown in Table \ref{tab:delete_example}.

\paragraph{Exchange} Exchange type low-quality data is characterized by the complete irrelevance between questions and their corresponding answers. In this approach, we swap answers among question-answer pairs to illustrate this form of low-quality data. Table \ref{tab:exchange_example} demonstrates examples of this exchange, where the first two samples involve swapping answers within the Medalpaca-flashcards dataset~\citep{han2023medalpaca}. The remaining two samples, drawn from PMC-Llama~\citep{wu2023pmc}, undergo a process where their rationales for answer choices are exchanged, and a different, incorrect choice is selected to ensure a clear distinction in the option outputs.

\begin{table}[htb]
\begin{tabular}{l p{3cm} p{4cm} p{4cm}}
\toprule
\textbf{Index}   & \textbf{Question}  & \textbf{Original Answer} & \textbf{Answer} 
\\ \midrule
11136 & '\#\#\#Question: What is the name of the agent used to treat pulmonary hypertension, and how does it work in relation to the Endothelin-1 (ET-1) receptor?'  & '\#\#\#Answer: The agent used to treat pulmonary hypertension is called Bosentan, and it works by blocking the Endothelin-1 (ET-1) receptor.'   & '\#\#\#Answer: In multiple myeloma, neoplastic plasma cells produce IL-6 which stimulates osteoclast activating factor.'  \\ \midrule
13747 & '\#\#\#Question: What specific substance do neoplastic plasma cells produce in multiple myeloma that stimulates osteoclast activating factor?'  & '\#\#\#Answer: In multiple myeloma, neoplastic plasma cells produce IL-6 which stimulates osteoclast activating factor.'  & '\#\#\#Answer: Permethrin is the anti-scabies/louse therapy that causes neuronal membrane depolarization.' 
\\ \midrule
4888  & '\#\#\#Question: Iodine deficiency control programme ?\textbackslash{}n\#\#\#Options:\textbackslash{}nA. Health education\textbackslash{}nB. Water testing\textbackslash{}nC. Foification of salt\textbackslash{}nD. None\textbackslash{}n'          & "\#\#\#Rationale: Ans. is 'c' i.e., Foification of salt Iodized salt (salt foified with iodine) is most economical, convenient and effective means of mass prophylaxis in endemic area.\textbackslash{}n\textbackslash{}n\#\#\#Answer: OPTION C IS CORRECT." 
& 
'\#\#\#Rationale: Fracture neck of talus results from forced dorsiflexion of the ankle. Typically this injury is sustained in an aircraft crash where the rubber bar is driven forcibly against the middle of the sole of the foot (ATOR\&;s fracture), resulting in forced dorsiflexion of the ankle, the neck being a weak area, gives way. Reference - Essential ohopaedics- Maheshwari -5th edn pg no 166.\textbackslash{}n\textbackslash{}n\#\#\#Answer: OPTION B IS CORRECT.' \\ \midrule
4066  & '\#\#\#Question: Ator fracture is -\textbackslash{}n\#\#\#Options:\textbackslash{}nA. Fracture neck of talus\textbackslash{}nB. Fracture scaphoid\textbackslash{}nC. Fracture calcaneum\textbackslash{}nD. Fracture 5th metatarsal\textbackslash{}n' & '\#\#\#Rationale: Fracture neck of talus results from forced dorsiflexion of the ankle. Typically this injury is sustained in an aircraft crash where the rubber bar is driven forcibly against the middle of the sole of the foot (ATOR\&;s fracture), resulting in forced dorsiflexion of the ankle, the neck being a weak area, gives way. Reference - Essential ohopaedics- Maheshwari -5th edn pg no 166.\textbackslash{}n\textbackslash{}n\#\#\#Answer: OPTION A IS CORRECT.' & "\#\#\#Rationale: Ans. is 'c' i.e., Foification of salt Iodized salt (salt foified with iodine) is most economical, convenient and effective means of mass prophylaxis in endemic area.\textbackslash{}n\textbackslash{}n\#\#\#Answer: OPTION D IS CORRECT." \\ \bottomrule

\end{tabular}
\caption{Synthetic low-quality examples of exchange type}
\label{tab:exchange_example}
\end{table}

\begin{table}[ht!]
\begin{tabular}{l p{5.5cm} p{4cm} p{2cm}}
\toprule
\textbf{Index}   & \textbf{Question}  & \textbf{Original Answer} & \textbf{Answer} 
\\ \midrule
12368 & '\#\#\#Question: What are some potential health risks associated with excess intake of Vitamin C?' 
& ‘\#\#\#Answer: Excess intake of Vitamin C can lead to a number of potential health risks, including the formation of calcium oxalate kidney stones, a condition known as calcium oxalate nephrolithiasis. This is because Vitamin C is broken down in the body to oxalate, a compound that can combine with calcium to form kidney stones. Other potential risks associated with high doses of Vitamin C include gastrointestinal issues such as diarrhea, nausea, and abdominal cramps. It is important to consume Vitamin C in moderation and to follow the recommended daily intake levels to avoid these potential health risks.' & '\#\#\#Answer: intake can lead a number of risks, the of calcium oxalate stones, a as calcium nephrolithiasis. because Vitamin C broken down in body to oxalate, form kidney Other risks with high of Vitamin include gastrointestinal issues as It to Vitamin C and to follow levels avoid these potential' 
\\ \midrule
7297  & '\#\#\#Question: Does liver damage in the rat induce hepatocyte stem cells from biliary epithelial cells?\textbackslash{}n\#\#\#Context: When rat hepatocyte regeneration after partial hepatectomy is blocked by 2-acetylaminofluorene, a proliferation of biliary epithelia sends out ductules into the parenchyma. The ability of these neoductules to act as a significant progenitor compartment for hepatocytes is in dispute. This study aims to resolve this question by varying the amount of 2-acetylaminofluorene administered. Rats were fed 2-acetylaminofluorene fr 6 days before and up to 7 days after partial hepatectomy was performed at a dose of either 2.5 (low) or 5 (high) mg/kg(-1)/day(-1). The response was monitored by the immunohistochemical expression of intermediate filaments and cytochrome P450 enzymes. No regeneration by mature hepatocytes occurred with either dose, and new ductules expressed the biliary cytokeratins 7, 8, 18, and 19 and, in addition, vimentin. At the high dose, hepatocytic differentiation was infrequent, whereas apoptosis and intestinal differentiation were common. At the low dose, almost all ductules differentiated into hepatocytes within 14 days of hepatectomy.' & 'Biliary epithelium is an effective and substantiative hepatocyte progenitor compartment under appropriate conditions.\textbackslash{}n\textbackslash{}n\#\#\#Answer: yes'                                                         & 'Biliary effective and substantiative hepatocyte under conditions.\textbackslash{}n\textbackslash{}n\#\#\#
Answer:'  
\\ \bottomrule
\end{tabular}
\caption{Synthetic low-quality examples of delete type}
\label{tab:delete_example}
\end{table}

\begin{table}[htb]
\begin{tabular}{l p{3.5cm} p{4cm} p{4cm}}
\toprule
\textbf{Index}   & \textbf{Question}  & \textbf{Original Answer} & \textbf{Answer} 
\\ \midrule
8002 & '\#\#\#Question: Which antibiotics are commonly prescribed to treat hepatic encephalopathy?'                  & "\#\#\#Answer: The antibiotics that are typically used to treat hepatic encephalopathy are rifaximin and neomycin. Hepatic encephalopathy is a neurological disorder that occurs as a result of liver disease. It is characterized by a range of symptoms such as confusion, disorientation, and impaired motor function. Antibiotics such as rifaximin and neomycin are used to reduce the levels of ammonia in the body, which can contribute to the development of hepatic encephalopathy. These antibiotics work by reducing the number of bacteria in the gut that produce ammonia. Rifaximin is preferred due to its lower risk of side effects compared to neomycin. However, the choice of antibiotic may depend on the individual patient's medical history and the severity of their condition." & '\#\#\#Answer: The antibiotics that are typically used to treat hepatic encephalopathy are rifaximin and neomycin. Hepatic encephalopathy is a neurological disorder that occurs as a result of liver disease. It is characterized by a range of symptoms such as confusion, disorientation, and impaired motor function. Antibiotics such as rifaximin and neomycin are used to reduce the levels of ammonia in the body, which can contribute to the development of hepatic encephalopathy. These antibiotics work by reducing the number of bacteria in the gut that produce ammonia. Rifaximin is preferred due to its lower risk of side effects compared to neomycin.' 
\\ \midrule
362  & '\#\#\#Question: Which of the following statement is TRUE about NPU (net protein utilization), used in protein quality assessment?\textbackslash{}n\#\#\#Options:\textbackslash{}nA. It is the ratio of energy from protein to total energy in diet\textbackslash{}nB. It is the ratio between nitrogen retained by the body and total nitrogen intake multiplied by 100\textbackslash{}nC. It is the amount of one amino acid per gram of a protein divided by the amount of same amino acid per gram of egg protein\textbackslash{}nD. If the NPU is high the amount of protein requirement in diet is high\textbackslash{}n' & "\#\#\#Rationale: Net Protein Utilization (NPU) is the ratio between nitrogen retained by the body and total nitrogen intake multiplied by 100. It is a product of digestibility coefficient and biological value divided by 100. It gives a more complete expression of protein quality than the amino acid score. In calculating protein quality, 1 gram of protein is assumed to be equivalent to 6.25 g of N. Ref: Park's Textbook of Preventive and Social medicine, 19th Edition, Page 503.\textbackslash{}n\textbackslash{}n\#\#\#Answer: OPTION B IS CORRECT." 
& "\#\#\#Rationale: Net Protein Utilization (NPU) is the ratio between nitrogen retained by the body and total nitrogen intake multiplied by 100. It is a product of digestibility coefficient and biological value divided by 100. It gives a more complete expression of protein quality than the amino acid score. In calculating protein quality, 1 gram of protein is assumed to be equivalent to 6.25 g of N. Ref: Park's Textbook of" 
\\ \bottomrule
\end{tabular}
\caption{Synthetic low-quality examples of cut type}
\label{tab:cut_example}
\end{table}
\end{document}